\documentclass[number]{ReportTemplate}
\usepackage{amssymb}
\usepackage{amsfonts}
\usepackage[fleqn]{amsmath}
\usepackage{mathtools}
\usepackage{amsthm}
\usepackage{bm}
\usepackage{url}
\usepackage{graphicx}
\usepackage{algorithmic}
\mathtoolsset{showonlyrefs,showmanualtags}

\newcommand{\OPT}{\bm x^{OPT}}

\begin{document}
\begin{frontmatter}
\title{On the approximation ability of evolutionary optimization with application to minimum set cover}
\author[nju]{Yang Yu}
\ead{yuy@lamda.nju.edu.cn}
\author[bham]{Xin Yao}
\ead{x.yao@cs.bham.ac.uk}
\infoauthor[nju]{Zhi-Hua Zhou}{\corref{cor1}}
\ead{zhouzh@nju.edu.cn}
\cortext[cor1]{Corresponding author}
\address[nju]{National Key Laboratory for Novel Software Technology, Nanjing University, Nanjing 210093, China}
\address[bham]{Center of Excellence for Research in Computational Intelligence and Applications, School of Computer Science, University of Birmingham, Birmingham B15 2TT, UK}

\begin{abstract}
Evolutionary algorithms (EAs) are heuristic algorithms inspired by natural evolution. They are often used to obtain satisficing solutions in practice. In this paper, we investigate a largely underexplored issue: the approximation performance of EAs in terms of how close the solution obtained is to an optimal solution. We study an EA framework named simple EA with isolated population (SEIP) that can be implemented as a single- or multi-objective EA. We analyze the approximation performance of SEIP using the partial ratio, which characterizes the approximation ratio that can be guaranteed. Specifically, we analyze SEIP using a set cover problem that is NP-hard. We find that in a simple configuration, SEIP efficiently achieves an $H_n$-approximation ratio, the asymptotic lower bound, for the unbounded set cover problem. We also find that SEIP efficiently achieves an $(H_k-\frac{k-1}{8k^9})$-approximation ratio, the currently best-achievable result, for the $k$-set cover problem. Moreover, for an instance class of the $k$-set cover problem, we disclose how SEIP, using either one-bit or bit-wise mutation, can overcome the difficulty that limits the greedy algorithm.
\end{abstract}

\begin{keyword}
Evolutionary algorithms \sep approximation algorithm  \sep approximation ratio \sep $k$-set cover \sep time complexity analysis
\end{keyword}

\end{frontmatter}

\section{Introduction}
Evolutionary algorithms (EAs) \cite{back:96} have been successfully applied to many fields and can achieve extraordinary performance in addressing some real-world hard problems, particularly NP-hard problems \cite{Higuchi.etal.tec99,Koza.etal.IS03,hornby.etal.space06,benkhelifa.etal.cec09}. To gain an understanding of the behavior of EAs, many theoretical studies have focused on the running time required to achieve exact optimal solutions \cite{he:yao:01,yu:zhou:aij08,neumann.witt.10,auger.doerr.11}. In practice, EAs are most commonly used to obtain \emph{satisficing} solutions, yet theoretical studies of the approximation ability of EAs have only emerged recently.

\citet{he_yao_03_cec} first studied conditions under which the \emph{wide-gap far distance} and the \emph{narrow-gap long distance} problems are hard to approximate using EAs. \citet{Giel_LNCS03} investigated a (1+1)-EA for a \emph{maximum matching} problem and found that the time taken time to find exact optimal solutions is exponential but is only $O(n^{2\lceil 1/\epsilon \rceil})$ for $(1+\epsilon)$-approximate solutions, which demonstrates the value of EAs as approximation algorithms.

Subsequently, further results on the approximation ability of EAs were reported. For the (1+1)-EA, the simplest type of EA, two classes of results have been obtained. On one hand, it was found that the (1+1)-EA has an arbitrarily poor approximation ratio for the \emph{minimum vertex cover} problem and thus also for the \emph{minimum set cover} problem \cite{Friedrich.etal.GECCO07,Oliveto.etal.TEC09}. On the other hand, it was also found that (1+1)-EA provides a \emph{polynomial-time randomized approximation scheme} for a subclass of the \emph{partition} problem \cite{witt.05}. Furthermore, for some subclasses of the minimum vertex cover problem for which the (1+1)-EA gets stuck, a multiple restart strategy allows the EA to recover an optimal solution with high probability \cite{Oliveto.etal.TEC09}. Another result for the (1+1)-EA is that it improves a $2$-approximation algorithm to a $(2-2/n)$-approximation on the minimum vertex cover problem \cite{Friedrich.etal.CEC07}. This implies that it might sometimes be useful as a post-optimizer.

Recent advances in multi-objective (usually bi-objective) EAs have shed light on the power of EAs as approximation optimizers. For a single-objective problem, \emph{multi-objective reformulation} introduces an auxiliary objective function for which a multi-objective EA is used as the optimizer. \citet{Scharnow.etal.PPSN02} first suggested that multi-objective reformulation could be superior to use of a single-objective EA. This was confirmed for various problems \cite{Neumann.Wegener.GECCO05,Neumann.Wegener.2007,Friedrich.etal.GECCO07,NEUMANN.GECCO08} by showing that while a single-objective EA could get stuck, multi-objective reformulation helps to solve the problems efficiently.

Regarding approximations, it has been shown that multi-objective EAs are effective for some NP-hard problems. \citet{Friedrich.etal.GECCO07} proved that a multi-objective EA achieves a $(\ln n)$-approximation ratio for the minimum set cover problem, and reaches the asymptotic lower bound in polynomial time. \citet{neumann.reichel.ppsn08} showed that multi-objective EAs achieve a $k$-approximation ratio for the \emph{minimum multicuts} problem in polynomial time.

In the present study, we investigate the approximation ability of EAs by introducing a framework called \emph{simple evolutionary algorithm with isolated population} (SEIP), which uses an isolation function to manage competition among solutions. By specifying the isolation function, SEIP can be implemented as a single- or multi-objective EA. Multi-objective EAs previously analyzed \cite{neumann_laumanns.LATIN06,Friedrich.etal.GECCO07,neumann.reichel.ppsn08} can be viewed as special cases of SEIP in term of the solutions maintained in the population. By analyzing the SEIP framework, we obtain a general characterization of EAs that guarantee approximation quality.

We then study the minimum set cover problem (MSCP), which is an NP-hard problem \cite{Feige.jacm98}. We prove that for the unbounded MSCP, a simple configuration of SEIP efficiently obtains an $H_k$-approximation ratio (where $H_k$ is the harmonic number of the cardinality of the largest set), the asymptotic lower bound \cite{Feige.jacm98}. For the minimum $k$-set cover problem, this approach efficiently yields an $(H_k-\frac{k-1}{8k^9})$-approximation ratio, the currently best-achievable quality~\cite{Hassin.Levin.SICOMP05}. Moreover, for a subclass of the minimum $k$-set cover problem, we demonstrate how SEIP, with either one-bit or bit-wise mutation, can overcome the difficulty that limits the greedy algorithm.

The remainder of the paper is organized as follows. After introducing preliminaries in Section 2, we describe SEIP in Section 3 and characterize its behavior for approximation in Section 4. We then analyze the approximation ratio achieved by SEIP for the MSCP in Section 5. In Section 6 we conclude the paper with a discussion of the advantages and limitations of the SEIP framework.

\section{Preliminaries}

We use bold small letters such as $\bm w, \bm x, \bm y,\bm z$ to represent vectors. We denote $[m]$ as the set $\{1,2,\ldots, m\}$ and $2^S$ as the power set of $S$, which consists of all subsets of $S$. We denote $H_n = \sum_{i=1}^{n} \frac{1}{i}$ for the $n$th harmonic number. Note that $H_n \sim \ln n$ since $\ln n \leq H_n \leq \ln n+1$.

In this paper, we consider minimization problems as follows.
\begin{Def}[Minimization problem]\label{def_problem}
Given an evaluation function $f$ and a set of feasibility constraints $\mathcal{C}$, find a solution $\bm x\in \{0,1\}^n$ that minimizes $f(\bm x)$ while satisfying constraints in $\mathcal{C}$. A problem instance can be specified by its parameters $(n, f, \mathcal{C})$.
\end{Def}

In the definition of the minimization problem, solutions are represented in binary vectors. When the aim of a minimization problem is to find a subset from a universal set, we can equivalently use a binary vector to represent a subset, where each element of the vector indicates the membership of a corresponding element of the universe set. For example, given a universal set $U=\{1,2,3,4,5\}$, its subset $S=\{1,3,5\}$ can be represented by a binary vector $\bm v=(1,0,1,0,1)$, and we define $U(\bm v)=S$. Considering the equivalence between sets and binary vectors, we apply set operators ($|\cdot|,\cap,\cup,-,\subseteq,\in$) to binary vectors when there is no confusion. For example, $|(1,0,1,0,1)|=3$, $(1,0,1,0,1)\cap (0,0,0,1,1) = (0,0,0,0,1)$, and $(1,0,1,0,1) - (0,0,0,1,1) = (1,0,1,0,0)$. We denote $\bm x^\emptyset = (0,\ldots,0)$ as the vector corresponding to the empty set.

We investigate the minimum set cover problem (MSCP), which is an NP-hard problem.
\begin{Def}[Minimum set cover problem (MSCP)]
Given a set of $n$ elements $U=[n]$ and a collection $C=\{S_1, S_2, \ldots, S_m\}$ of $m$ nonempty subsets of $U$, where each $S_i$ is associated with a positive cost $w(S_i)$, find a subset $X^*\subseteq C$ such that $\sum_{S\in X^*} w(S)$ is minimized with respect to $\mathop{\cup}\limits_{S\in X^*} S = U$.
\end{Def}

Using binary vector representation, we denote an instance of the MSCP by its parameters $(n,\bm w,C, U)$, where $|C|=m$ and $\bm w$ is the cost vector. The MSCP involves finding a vector $\OPT$, which is equivalent to its set representation $X^*$, by solving a constrained optimization problem

\begin{align}
\OPT = \mathop{\arg\min}_{\bm x\in \{0,1\}^m} \bm w \cdot \bm x \quad \\
\text{s.t.} \quad \bigcup_{S\in C(\bm x)} S = U ,
\end{align}

where $\bm w\cdot\bm x$ is the inner product between vectors $\bm w$ and $\bm x$, and $C(\bm x)$ denotes a set consisting of elements in the collection $C$ that are indicated by binary vector $\bm x$.

\begin{Def}[Minimum $k$-set cover problem]
An MSCP $(n,\bm w, C, U)$ is a $k$-set cover problem if, for some constant $k$, it holds that $|S|\leq k$ for all $S\in C$, denoted as $(n,\bm w, C, U,k)$.
\end{Def}

A solution is called \emph{feasible} if it satisfies all the constraints in $\mathcal{C}$; otherwise, it is called an \emph{infeasible} solution, or a \emph{partial} solution in this paper. Here, we assume that the evaluation function $f$ is defined on the full input space, that is, it evaluates all possible solutions regardless of their feasibility. For example, if we evaluate a solution $\bm x$ of the MSCP using $\bm w\cdot\bm x$, this evaluation function can be calculated for any solution.

Given a minimization problem, we denote $\OPT$ as an optimal solution of the problem, and $OPT=f(\OPT)$. For a feasible solution $\bm x$, we regard the ratio $\frac{f(\bm x)}{OPT}$ as its \emph{approximation ratio}. If the approximation ratio of a feasible solution is upper-bounded by some value $r$, that is,
\begin{align}
1\leq \frac{f(\bm x)}{OPT} \leq r,
\end{align}
the solution is called an \emph{$r$-approximate solution}. An algorithm that guarantees to find an $r$-approximate solution for an arbitrary problem instance in polynomial time is an \emph{$r$-approximation algorithm}.

The greedy algorithm \cite{cheatal.mos79} described here as Algorithm \ref{algo_greedy} is the most well-known approximation algorithm for the MSCP. This greedy algorithm consists of a sequence of steps. The cost of a candidate set is defined as its weight divided by the number of its elements that have not been covered yet (i.e., the quantity $r_S$ in line 3). The algorithm then picks the candidate set with the smallest cost for the solution (line 4) and marks the newly covered elements (line 6). This simple algorithm yields an $H_n$-approximation ratio or, more exactly, $H_k$, where $k$ is the cardinality of the largest set \cite{cheatal.mos79}. The key to the proof of the approximation ratio is the definition of the $price$ of elements, as in line 5. The $price$ of an element equals the cost of the set that first covers it; therefore, the total $price$ of all elements equals the total cost of the solution. Furthermore, it should be noted that when an element is covered by a set with the lowest cost, it would also be covered by one set of an optimal solution but with a higher cost. Therefore the $price$ of the element is upper-bounded by the optimal cost and hence the approximation ratio is upper-bounded. For a detailed proof please refer to \citet{cheatal.mos79}.

\begin{Algo}[Greedy algorithm \cite{cheatal.mos79}]\label{algo_greedy}\small
Given a minimum $k$-set cover problem $(n, \bm w, C, U,k)$, the greedy algorithm consists of the following steps:
\begin{algorithmic}[1]
\STATE $X\gets \emptyset; R\gets \emptyset$
\WHILE{$R\neq U$}
\STATE $\forall S\in C : |S -  R|>0$, let $r_S \gets \frac{ w(S)}{|S -  R|}$
\STATE $\hat{S} \gets \mathop{\arg\min}\limits_S r_S$
\STATE let $price(e) \gets r_{\hat{S}}$ for all $e\in \hat{S}  -  R$
\STATE let $R \gets R \cup \hat{S}$, and $X \gets X \cup \{\hat{S}\}$.
\ENDWHILE
\RETURN $X$
\end{algorithmic}
\end{Algo}

Several studies have shown that the approximation ratio of the MSCP is lower-bounded by $\Omega(\ln n)$ unless $P=NP$ that is unlikely \cite{raz.safra.stoc97,slavik.jalgo97,Feige.jacm98,Alon.etal.06}. Therefore, the greedy algorithm achieves the asymptotic lower bound of the approximation ratio for the MSCP.

Although the $H_n$-approximation ratio is asymptotically tight for the unbounded MSCP, a better approximation ratio can be achieved for the minimum $k$-set cover problem, where $k$ is a constant. It has been proved that for the unweighted minimum $k$-set cover problem, an $(H_k-\frac{1}{2})$-approximation ratio can be achieved \cite{duh.furer.stoc97} and, if $k\geq 4$, an improved ratio $(H_k-\frac{196}{390})$ can be achieved \cite{Levin08}. For the weighted minimum $k$-set cover problem, a greedy-algorithm-with-withdrawals (GAWW) was presented and achieved an $H_k - \frac{k-1}{8k^9}$-approximation ratio \cite{Hassin.Levin.SICOMP05}.

The GAWW algorithm presented here as Algorithm \ref{algo_gaww} is a modification of the greedy algorithm. In every iteration, the algorithm chooses between taking a greedy step, which is the same as in the greedy algorithm, and a withdrawal step, which replaces a set in the current solution with at most $k$ candidate sets. It evaluates the cost of candidate sets as in the greedy algorithm, and also evaluates the benefit of the withdrawal (calculated in lines 4 and 5). When the benefit of the withdrawal step is not large enough according to the criterion in line 6, the algorithm takes the greedy step, and otherwise takes the withdrawal step. To prove the approximation ratio, the $price$ of elements is defined similarly in line 7 for the greedy step and line 10 for the withdrawal step, which is used later in the proofs for this paper.

\begin{Algo}[GAWW \cite{Hassin.Levin.SICOMP05}]\label{algo_gaww}\small
Given a minimum $k$-set cover problem $(n, \bm w, C, U,k)$, GAWW consists of the following steps.
    \begin{algorithmic}[1]
    \STATE $X\gets \emptyset; R\gets \emptyset; \alpha_k \gets 1-\frac{1}{k^3}$
    \WHILE{$R\neq U$}
        \STATE $\forall S\in C : |S -  R|>0$, let $r_S \gets \frac{ w(S)}{|S -  R|}$
        \STATE $\forall S\in X, Q\subseteq C: |Q|\leq k \wedge |\cup_{S'\in Q} S'  -  R| >0 $, let $r_{(S,Q)} \gets \frac{\sum_{S'\in Q} c_{S'} - c_S}{|\cup_{S'\in Q} S'  -  R|}$
        \STATE $\hat{S} \gets \mathop{\arg\min}\limits_S r_{\hat{S}}$, and $(\tilde{S},\tilde{Q})\gets \mathop{\arg\min}\limits_{\mathclap{(S,Q):|Q|\leq k}} r_{(S,Q)}$
        \COMMENT {choose the minimal number of sets in cases of ties}
        \IF[greedy step]{$r_{\hat{S}}\cdot \alpha_k \leq r_{(\tilde{S},\tilde{Q})}$}
            \STATE let $price(e) \gets r_{\hat{S}}$ for all $e\in \hat{S}  -  R$
            \STATE let $R \gets R \cup \hat{S}$, and $X \gets X \cup \{\hat{S}\}$.
        \ELSE [i.e., $r_{(\tilde{S},\tilde{Q})} < r_{\hat{S}}\cdot\alpha_k$, withdrawal step]
            \STATE let $price(e) \gets r_{(\tilde{S},\tilde{Q})}$ for all $e\in \cup_{S\in Q} S  -  R$
            \STATE let $R \gets \cup_{S\in Q} S \cup R$, and $X \gets X \cup Q  - \{\tilde{S}\}$
        \ENDIF
    \ENDWHILE
    \RETURN $X$
    \end{algorithmic}
\end{Algo}

The (1+1)-EA is the simplest EA implementation, as described in Algorithm \ref{algo_1+1}. Starting from a solution generated uniformly at random, the (1+1)-EA repeatedly generates a new solution from the current one using a mutation operator, and the current solution is replaced if the new solution has better (or equal) fitness.

\begin{Algo}[(1+1)-EA]\label{algo_1+1}\small
Given a minimization problem $(n, f,\mathcal{C})$, each solution is encoded as a binary vector of length $m$ and the (1+1)-EA-minimizing $f$ consists of the following steps.
    \begin{algorithmic}[1]
        \STATE $\bm x \gets \text{a solution generated uniformly at random}$
        \WHILE {$\mathtt{stop}\neq\mathtt{true}$}
            \STATE Mutate $\bm x$ to generate $\bm x'$
            \IF {$x$ is feasible and $f(\bm x')\leq f(\bm x)$}
                \STATE $\bm x\gets \bm x'$
            \ENDIF
        \ENDWHILE
        \RETURN $\bm x$
    \end{algorithmic}
\end{Algo}

Two mutation operators are commonly used to implement the ``mutate'' step in line 3:\\
\indent \textbf{One-bit mutation:} Flip one randomly selected bit position of $\bm x$ to generate $\bm x'$.\\
\indent \textbf{Bit-wise mutation:} Flip each bit of $\bm x$ with probability $\frac{1}{m}$ to generate $\bm x'$.\\

It has been shown that the (1+1)-EA has an arbitrarily poor approximation ratio for the MSCP \cite{Friedrich.etal.GECCO07}. \citet{Laumanns.etal.PPSN02} used a multi-objective reformulation with a multi-objective EA named SEMO to achieve a $(\ln n)$-approximation ratio. The SEMO algorithm is described in Algorithm \ref{algo_semo}, where two objectives are presented as $(f_1, f_2)$. To apply SEMO to the MSCP, let $f_1$ evaluate the cost of the solution and $f_2$ evaluate the number of uncovered elements. Thus, SEMO minimizes the cost and the number of uncovered elements of solutions simultaneously. A notable difference between SEMO and (1+1)-EA is that SEMO uses a non-dominance relationship, implemented using the $\mathtt{dominate}$ function in SEMO. The population of SEMO maintains non-dominant solutions, that is, no solution is superior to another for both of the two objectives.

\begin{Algo}[SEMO \cite{Laumanns.etal.PPSN02}]\label{algo_semo}\small
Given a two-objective minimization problem $(n, (f_1, f_2))$, each solution is encoded as a binary vector of length $m$. SEMO minimization of $(f_1, f_2)$ consists of the following steps.

    \begin{algorithmic}[1]
        \STATE $P \gets \{\bm x^\emptyset=(0,0,\ldots,0)\}$
        \WHILE {$\mathtt{stop}\neq\mathtt{true}$}
            \STATE Choose $\bm x\in P$ uniformly at random
            \STATE \label{algo:line:mutation} Mutate $\bm x$ to generate $\bm x'$
            \IF {$\forall \bm x\in P: \mathtt{dominate}(\bm x,\bm x')=\mathtt{false}$}
                \STATE $Q \gets \{\bm x\in P \mid \mathtt{dominate}(\bm x',\bm x)=\mathtt{true}\}$
                \STATE $P\gets P\cup \{\bm x'\} - Q$
            \ENDIF
        \ENDWHILE
        \RETURN $P$
    \end{algorithmic}
    where the $\mathtt{dominate}$ function of two solutions is defined such that $\mathtt{dominate}(\bm x,\bm y)$ is $\mathtt{true}$ if any one of the following three rules is satisfied:\\
    1) $f_1(\bm x) < f_1(\bm y)$ and $f_2(\bm x) <= f_2(\bm y)$ \\
    2) $f_1(\bm x) <= f_1(\bm y)$ and $f_2(\bm x) < f_2(\bm y)$ \\
    3) $f_1(\bm x)=f_1(\bm y)$ and $f_2(\bm x) = f_2(\bm y)$ and $|\bm x|<|\bm y|$\\
    and $\mathtt{dominate}(\bm x,\bm y)$ is $\mathtt{false}$ otherwise.
\end{Algo}

\section{SEIP}

The SEIP framework is depicted in Algorithm \ref{algo_gsel}. It uses an \emph{isolation function} $\mu$ to isolate solutions. For some integer $q$, the function $\mu$ maps a solution to a subset of $[q]$. If and only if two solutions $\bm x_1$ and $\bm x_2$ are mapped to subsets with the same cardinality, that is, $|\mu(\bm x_1)|=|\mu(\bm x_2)|$, the two solutions compete with each other. In that case, we say the two solutions are in the same \emph{isolation}, and there are at most $q+1$ isolations since the subsets of $[q]$ have $q+1$ different cardinalities.

\begin{Algo}[SEIP]\label{algo_gsel}\small
Given a minimization problem $(n, f, \mathcal{C})$, an isolation function $\mu$ encodes each solution as a binary vector of length $m$. SEIP minimization of $f$ with respect to constraint $\mathcal{C}$ consists of the following steps.

    \begin{algorithmic}[1]
        \STATE $P \gets \{\bm x^\emptyset=(0,0,\ldots,0)\}$
        \WHILE {$\mathtt{stop}\neq\mathtt{true}$}
            \STATE Choose $\bm x\in P$ uniformly at random
            \STATE Mutate $\bm x$ to generate $\bm x'$
            \IF {$\forall \bm x\in P: \mathtt{superior}(\bm x,\bm x')=\mathtt{false}$}
                \STATE $Q \gets \{\bm x\in P \mid \mathtt{superior}(\bm x',\bm x)=\mathtt{true}\}$
                \STATE $P\gets P\cup \{\bm x'\} - Q$
            \ENDIF
        \ENDWHILE
        \RETURN the best feasible solution in $P$
    \end{algorithmic}
    where the $\mathtt{superior}$ function of two solutions determines whether one solution is superior to the other. This is defined as follows: $\mathtt{superior}(\bm x,\bm y)$ is $\mathtt{true}$ if both of the following rules are satisfied:\\
    1) $|\mu(\bm x)|=|\mu(\bm y)|$ \\
    2) $f(\bm x)<f(\bm y)$, or $f(\bm x)=f(\bm y)$ but $|\bm x|<|\bm y|$\\
    and $\mathtt{superior}(\bm x,\bm y)$ is $\mathtt{false}$ otherwise.
\end{Algo}

When the isolation function puts all solutions in an isolation for a particular instance, it degrades to the (1+1)-EA. The isolation function can also be configured to simulate the dominance relationship of multi-objective EAs such as SEMO/GSEMO \cite{Laumanns.etal.PPSN02} and DEMO \cite{neumann.reichel.ppsn08}. If we are dealing with $k$-objective optimization with discrete objective values, a simple approach is to use one of the objective functions, say $f_1$, as the fitness function and use the combination of the values of the remaining $k-1$ objective functions (say $f_2, \ldots, f_k$) as the isolation functions. Thus, two solutions compete (for $f_1$) only when they share the same objective values for $f_2,\ldots, f_k$. This simulation shows that all the non-dominant solutions of a multi-objective EA are also kept in the SEIP population, and if a non-dominant solution does not reside in the SEIP population, there must be another solution that dominates it. Hence, SEIP can be viewed as a generalization of multi-objective EAs in terms of the solutions retained. This simulation also reveals that SEIP retains more solutions than a multi-objective EA using the dominance relationship. This, on one hand, SEIP takes more time to manipulate a larger population than a multi-objective EA does, which could be overcome by defining an isolation function that aggregates nearby solutions, as has been done for DEMO \cite{neumann.reichel.ppsn08}. On the other hand, SEIP has more opportunities available to find a better approximation solution, since the relationship ``$\bm a$ dominates $\bm b$'' does not imply that $\bm a$ definitely leads to a better approximation solution than $\bm b$.

Taking the MSCP $(n,\bm w, C, U)$ as an example, we can use the fitness function $f(\bm x)=\bm w\cdot \bm x$, which is the sum of costs of the selected sets. For the isolation function, we can use $\mu(\bm x)$ as $\emptyset$ if $\bm x$ is feasible and $\{1\}$ if $\bm x$ is infeasible, which isolates the feasible from the infeasible solutions (and thus $q=1$); we can also use the isolation function $\mu(\bm x)=\cup_{S\in \bm x(C)}S$, and thus the solutions compete only when they cover the same number of elements (and thus $q=n$).

The mutation operator can use either one-bit or bit-wise mutation. Usually, one-bit mutation is considered for local searches, while bit-wise mutation is suitable for global searches as it has a positive probability for producing any solution. We denote SEIP with one-bit mutation as LSEIP and SEIP with bit-wise mutation as GSEIP, where ``L'' and ``G'' denote ``local'' and ``global'', respectively.

For convenience, SEIP is set to start from solution $\bm x^\emptyset$ rather than from a random solution, as commonly done for other EAs. Under the condition that any solution will have a better fitness if any 1 bit is turned to 0, we can bound the difference between random initialization and starting from $\bm x^\emptyset$. From a random solution, SEIP takes at most $O(qm\ln m)$ expected steps to find $\bm x^\emptyset$ according to the following argument. Suppose the worst case whereby random initialization generates a solution with all 1 bits; according to the fitness function condition, finding $\bm x^\emptyset$ is equivalent to solving the OneMax problem using a randomized local search and the (1+1)-EA, which takes $O(m\ln m)$ steps for both LSEIP and GSEIP \cite{droste.etal.ecj98}. Furthermore, note that there can be at most $q$ solutions in the population, and thus it costs SEIP $q$ expected steps to choose one particular solution from the population.

The $\mathtt{stop}$ criterion is not described in the definition of SEIP, since EAs are usually used as anytime algorithms in practice. We now analyze the approximation ratio and the corresponding computational complexity of SEIP.

\section{General approximation behavior of SEIP}

For minimization problems, we consider \emph{linearly additive} isolation functions. $\mu$ is a linearly additive isolation function if, for some integer $q$,
\begin{align}
    \begin{cases}
    \mu(\bm x)=[q],& \text{for all feasible solutions $\bm x$},\\
    \mu(\bm x\cup\bm y) = \mu(\bm x)\cup\mu(\bm y), & \text{for all solutions $\bm x$ and $\bm y$}.
    \end{cases}
\end{align}

The quality of a feasible solution is measured in terms of the approximation ratio. To measure the quality of a partial (infeasible) solution, we define a \emph{partial reference function} and \emph{partial ratio} as follows.

\begin{Def}[Partial reference function]\label{def_partialref}
Given a set $[q]$ and a value $v$, a function $\mathcal{L}_{[q],v}:2^{[q]}\to \mathbb{R}$ is a partial reference function if\\
    1) $\mathcal{L}_{[q],v}([q]) = v$,\\
    2) $\mathcal{L}_{[q],v}(R_1) = \mathcal{L}_{[q],v}(R_2)$ for all $R_1, R_2\subseteq [q]$ such that $|R_1|=|R_2|$.
\end{Def}
For a minimization problem with optimal cost $OPT$ and an isolation function $\mu$ mapping feasible solutions to the set $[q]$, we denote a partial reference function with respect to $[q]$ and $OPT$ as $\mathcal{L}_{[q],OPT}$. When the problem and the isolation function are clear, we omit the subscripts and simply denote the partial reference function as $\mathcal{L}$.

\begin{Def}[Partial ratio]\label{def_partialratio}
Given a minimization problem $(n, f, \mathcal{C})$ and an isolation function $\mu$, the \emph{partial ratio} of a (partial) solution $\bm x$ with respect to a corresponding partial reference function $\mathcal{L}$ is
    \begin{align}
       \textsf{p-ratio}(\bm x)=\frac{f(\bm x)}{\mathcal{L}(\mu(\bm x))},
    \end{align}
    and the \emph{conditional partial ratio} of $\bm y$ conditioned on $\bm x$ is
    \begin{align}
        \textsf{p-ratio}(\bm x\mid\bm y)=\frac{f( \bm y\mid\bm x)}{\mathcal{L}(\mu(\bm y)\mid \mu(\bm x))},
    \end{align}
    where $f(\bm y\mid \bm x) = f(\bm x\cup\bm y) - f(\bm x)$ and $\mathcal{L}(\mu(\bm y)\mid \mu(\bm x)) = \mathcal{L}(\mu(\bm y) \cup \mu(\bm x))-\mathcal{L}(\mu(\bm x))$.
\end{Def}

The partial ratio is an extension of the approximation ratio. Note that the partial ratio for a feasible solution equals its approximation ratio. We have two properties of the partial ratio. One is that it is non-increasing in SEIP, as stated in Lemma \ref{lem_noincrease}, and the other is its decomposability, as stated in Lemma \ref{lem_prdecomp}.

\begin{Lem}\label{lem_noincrease}
Given a minimization problem $(n, f, \mathcal{C})$ and an isolation function $\mu$, if SEIP has generated an offspring $\bm x$ with partial ratio $p$  with respect to a corresponding partial reference function $\mathcal{L}$, then there is a solution $\bm y$ in the population such that $|\mu(\bm y)|=|\mu(\bm x)|$, and the partial ratio of $\bm y$ is at most $p$.
\end{Lem}
\begin{myproof}
$\bm x$ is put into the population after it is generated; otherwise there is another solution $\bm x'$ with $|\mu(\bm x')|=|\mu(\bm x)|$ and $f(\bm x')\leq f(\bm x)$, and in this case let $\bm x=\bm x'$. The lemma is proved since $\mathcal{L}(\bm x)=\mathcal{L}(\bm y)$ and by the \texttt{superior} function the cost is non-increasing.
\end{myproof}

From Lemma \ref{lem_noincrease}, we know that the partial ratio in each isolation remains non-increasing. Since SEIP repeatedly tries to generate solutions in each isolation, SEIP can be considered as optimizing the partial ratio in each isolation.

\begin{Lem}\label{lem_prdecomp}
Given a minimization problem $(n, f, \mathcal{C})$ and an isolation function $\mu$, for three (partial) solutions $\bm x,\bm y$ and $\bm z$ such that $\bm z=\bm x\cup \bm y$, we have $$\textsf{p-ratio}(\bm z)\leq \max\{\textsf{p-ratio}(\bm x),\textsf{p-ratio}(\bm y\mid\bm x)\},$$ with respect to a corresponding partial reference function $\mathcal{L}$.
\end{Lem}
\begin{myproof}
Since $\bm z=\bm x\cup \bm y$, we have, by definition,
        \begin{align}
                f(\bm z) = f(\bm x) + f(\bm y\mid \bm x),
        \end{align}
        and
        \begin{align}
                \mu(\bm z) = \mu(\bm x\cup\bm y) = \mu(\bm x) \cup \mu(\bm y).
        \end{align}
        Thus, we have
        \begin{align}
        \textsf{p-ratio}(\bm z)=\frac{f(\bm z)}{\mathcal{L}(\mu(\bm z))} & =  \frac{f(\bm x)+f(\bm y\mid\bm x)}{\mathcal{L}(\mu(\bm x)\cup\mu(\bm y))} \\
        & =\frac{f(\bm x)+f(\bm y\mid\bm x)}{\mathcal{L}(\mu(\bm x))+\mathcal{L}(\mu(\bm y)\mid \mu(\bm x))}\\
        & \leq  \max\left\{
        \frac{f( \bm x)}{\mathcal{L}(\mu(\bm x))},
        \frac{f( \bm y\mid\bm x)}{\mathcal{L}(\mu(\bm y)\mid \mu(\bm x))}
        \right\} \\
        &= \max\{\textsf{p-ratio}(\bm x), \textsf{p-ratio}(\bm y\mid\bm x)\}.
        \end{align}
\end{myproof}

Lemma \ref{lem_prdecomp} reveals that the partial ratio for a solution is related to the conditional partial ratio of a building block. This can be considered as the way in which SEIP optimizes the partial ratio in each isolation, that is, by optimizing the conditional partial ratio of each building block partial solution. We then have the following theorem.

\begin{Them}\label{them_maxblock}
Given a minimization problem $(n, f, \mathcal{C})$ and an isolation function $\mu$ mapping to subsets of $[q]$, assume that every solution is encoded in an $m$-length binary vector. For some constant $r\geq 1$ with respect to a corresponding partial reference function $\mathcal{L}$, if
        \begin{enumerate}
                           \item $\textsf{p-ratio}(\bm x^\emptyset)\leq r$,
                           \item for every partial solution $\bm x$ such that $\textsf{p-ratio}(\bm x)\leq r$, SEIP takes $\bm x$ as the parent solution and generates an offspring partial solution $\bm y$ such that $\mu(\bm x) \subset \mu(\bm x\cup \bm y)$ and $\textsf{p-ratio}(\bm y\mid\bm x)\leq r$ in polynomial time in $q$ and $m$,
                         \end{enumerate}
        then SEIP finds an $r$-approximate solution in polynomial time in $q$ and $m$.
\end{Them}
\begin{myproof}
        Starting from $\bm x^\emptyset$, we can find a sequence of partial solutions $\bm x_1,\bm x_2\ldots, \bm x_\ell$, such that
        \begin{align}
                \bm x = \cup_{i=1}^\ell \bm x_i \text{ is feasible,}
        \end{align}
        and that
        \begin{align}
                \forall i = 1 \ldots \ell:
                \textsf{p-ratio}(\bm x_i \mid \bm x^\emptyset \cup_{j=1}^{i-1} \bm x_j )\leq r,
        \end{align}
        because of the conditions. Note that when a partial solution is added to the solution, the offspring solution is in a different isolation to the parent solution. Since the isolation function is linearly additive, the length of the sequence $\ell$ cannot be greater than the number of isolations $q+1$.

Let $t$ be the time expected for SEIP to generate a partial solution $\bm x_i$ in the sequence from its parent solution, which is polynomial to $q$ and $m$ by the condition. It takes at most $O(q)$ expected steps for SEIP to pick the parent, since there are at most $q+1$ solutions in the population. Therefore, the total time to reach a feasible solution $\bm x$ is $O(t\cdot q \cdot \ell)$, that is, $O(t\cdot q^2)$, which is still polynomial in $q$ and $m$.

By Lemma \ref{lem_prdecomp}, since the feasible solution $\bm x$ is composed of $\bm x^\emptyset$ and partial solutions $\bm x_1,\bm x_2\ldots, \bm x_m$, the approximation ratio for $\bm x$ is at most as large as the maximum conditional partial ratio for the partial solutions, $r$.
\end{myproof}

Theorem \ref{them_maxblock} reveals how SEIP can work to achieve an approximate solution. Starting from the empty set, SEIP uses its mutation operator to generate solutions in all isolations, and finally generates a feasible solution. During the process, SEIP repeatedly optimizes the partial ratio in each isolation by finding partial solutions with better conditional partial ratios. Since the feasible solution can be viewed as a composition of a sequence of partial solutions from the empty set, the approximation ratio is related to the conditional partial ratio of each partial solution.

In Theorem \ref{them_maxblock}, the approximation ratio is upper-bounded by the maximum conditional partial ratio, while some building-block partial solutions may have lower conditional partial ratios but are not utilized. Moreover, in Theorem \ref{them_maxblock} we restrict SEIP to append partial solutions, while GSEIP can also remove partial solutions using bit-wise mutation. The approximation ratio can have a tighter bound if we consider these two issues. Applying the same principle as for Theorem \ref{them_maxblock}, we present a theorem for GSEIP in particular that leads to a tighter approximation ratio.

\begin{Def}[Non-negligible path]\label{def_isopath}
Given a minimization problem $(n, f, \mathcal{C})$ and an isolation function $\mu$ mapping to subsets of $[q]$, assume that every solution is encoded in an $m$-length binary vector. A set of solutions $N$ is a \emph{non-negligible path} with \emph{ratios} $\{r_i\}_{i=0}^{q-1}$ and \emph{gap} $c$ if $\bm x^\emptyset\in N$ and, for every solution $\bm x \in N$, there exists a solution $\bm x' = (\bm x\cup \bm y^+ - \bm y^-) \in N$, where the pair of solutions $(\bm y^+, \bm y^-)$ satisfies
        \begin{enumerate}
          \item $1\leq |\bm y^+| + |\bm y^-|\leq c$,
          \item $f(\bm y^+ - \bm y^-\mid \bm x)\leq r_{|\mu(\bm x)|}\cdot OPT$,
          \item if $|\mu(\bm x)|<q$, $|\mu(\bm x\cup \bm y^+ - \bm y^-)| > |\mu(\bm x)|$.
        \end{enumerate}
\end{Def}

\begin{Them}\label{them_gsel}
Given a minimization problem $(n, f, \mathcal{C})$ and an isolation function $\mu$ mapping to subsets of $[q]$, assume that every solution is encoded in an $m$-length binary vector. If there exists a non-negligible path with ratios $\{r_i\}_{i=0}^{q-1}$ and gap $c$, then GSEIP finds an $(\sum_{i=0}^{q-1} r_i)$-approximate solution in expected time $O(q^2 m^c)$.
\end{Them}
\begin{myproof}
We prove the theorem by tracking the process of GSEIP over the non-negligible path. We denote by $\bm x^{cur}$ the solution we want to operate on. Initially, $\bm x^{cur}=\bm x^\emptyset$, and thus $|\mu(\bm x^{cur})|=0$.

GSEIP takes at most $O(q)$ expected steps to operate on $\bm x^{cur}$, since there are at most $q+1$ solutions in the population and GSEIP selects one to operate on in each step.

According to the definition of the non-negligible path, there exists a pair of solutions $( \bm y^+,\bm y^-)$ with respect to $\bm x^{cur}$ such that $1\leq |\bm y^+|+|\bm y^-|\leq c$. We denote $\bm x'=\bm x^{cur}\cup \bm y^+ - \bm y^-$. The probability that the mutation operator generates solution $\bm x'$ is at least $(\frac{1}{m})^c(\frac{m-1}{m})^{l-c}$, which implies $O(m^c)$ expected steps.

According to the definition of the non-negligible path, suppose that $|\mu(\bm x^{cur})|=i$; we also have $ f(\bm y^+ - \bm y^- \mid \bm x^{cur})\leq r_i\cdot OPT$. Note that $f(\bm x')$ can be decomposed recursively, and thus according to the theorem conditions, we have
         \begin{align}
                f(\bm x') & = f(\bm y^+ - \bm y^- \mid \bm x^{cur}) + f(\bm x^{cur}) \\
                & = r_{i} \cdot OPT+ f(\bm x^{cur})  = ... \\
                & = \sum_{j=0}^{i} r_j \cdot OPT + f(\bm x^\emptyset) = \sum_{j=0}^{i} r_j \cdot OPT.
         \end{align}
         Let $\mathcal{L}$ be a corresponding partial reference function. Thus, $\textsf{p-ratio}(\bm x') = \frac{\sum_{j=0}^{i} r_j\cdot OPT}{\mathcal{L}(\mu(\bm x'))}$.

         Given $|\mu(\bm x^{cur})|=i$, again according to the definition of the non-negligible path, we have $|\mu(\bm x')|> |\mu(\bm x)|$. Then we store solution $\bm x'$ in the population; otherwise there exists another solution $\bm x''$ with $|\mu(\bm x'')|=|\mu(\bm x')|$ and $\bm x''$ has a smaller partial ratio than $\bm x'$ by Lemma \ref{lem_noincrease} when we substitute $\bm x''$ for $\bm x'$.

         Now let $\bm x^{cur}$ be $\bm x'$. We have $\textsf{p-ratio}(\bm x^{cur}) \leq \frac{\sum_{j=0}^{ |\mu(\bm x^{cur})|-1 } r_j\cdot OPT}{\mathcal{L}(\mu(\bm x'))}$.

        After at most $q$ iterations of the above update of $\bm x^{cur}$, we have $|\mu(\bm x^{cur})|=q$, which means $\bm x^{cur}$ is feasible. Thus, the partial ratio of $\bm x^{cur}$, $\textsf{p-ratio}(\bm x^{cur}) =  \frac{\sum_{j=0}^{q-1} r_j\cdot OPT}{\mathcal{L}(\mu(\bm x^{cur}))}= \frac{\sum_{j=0}^{q-1} r_j\cdot OPT}{OPT} = \sum_{j=0}^{q-1} r_j$, is its approximation ratio.

         Thus, at most $q$ jumps are needed to reach a feasible solution, each takes $O(m^c)$ expected steps for operation on a particular solution, and it takes $O(q)$ expected steps to choose the particular solution. Overall, it takes $O(q^2m^c)$ expected steps to achieve the feasible solution.
\end{myproof}

Using Theorem \ref{them_gsel} to prove the approximation ratio of GSEIP for a specific problem, we need to find a non-negligible path and then calculate the conditional evaluation function for each jump on the path. One way of finding a non-negligible path for a problem is to follow an existing algorithm for the problem. This will lead to a proof that the EA can achieve the same approximation ratio by simulating the existing algorithm. Similar ideas have been used to confirm that EAs can simulate dynamic programming \cite{Doerr.09}. In addition, note that the concrete form of the partial reference function is not required in the theorem.

\section{SEIP for the MSCP}

To apply SEIP to the MSCP, we use the fitness function
 $$
   f(\bm x)=\bm w\cdot \bm x,
$$
which has the objective of minimizing the total weight. For a solution $\bm x$, we denote $R(\bm x) = \cup_{S\in \bm x(C)} S$, that is, $R(\bm x)$ is the set of elements covered by $\bm x$. We use the isolation function
$$
  \mu(\bm x) = R(\bm x),
$$
which, owing to the effect of the isolation function, makes two solutions compete only when they cover the same number of elements. We could regard a partial reference function $\mathcal{L}$ of $\bm x$ to be the minimum price that optimal solutions pay for covering the same number of elements covered by $\bm x$, although it is not necessary to calculate the partial reference function.

Instead of directly assessing a minimum $k$-set cover problem $(n, \bm w, C, U,k)$, we analyze EAs for the \emph{extended input} $(n, \bm w', C', U, k)$ \cite{Hassin.Levin.SICOMP05}. The original problem is extended by first taking a closure of $C$ under the subset operation, that is, $C' = C \cup \{2^S \mid \forall S\in C\}$, and the weight vector $\bm w'$ is extended accordingly by $ w'(S) = \min \{ w'(S_1), w'(S_2),\ldots, w'(S_j)\}$ if $S \subseteq  S_1,S_2,\ldots,S_j$. Then if an optimal solution contains a set with less than $k$ elements, we construct a new problem instance in which to $U$ are added a minimum number of dummy elements such that all sets of the optimal solution are filled to be $k$-sets using dummy elements while keeping their weights unchanged. Therefore, the extended problem has an optimal solution containing $k$-sets. Analysis on the extended input leads to the same result as for the original problem, as shown in Lemma \ref{lem_extension}. The lemma is derived from Lemmas 2 and 3 of \cite{Hassin.Levin.SICOMP05}.

\begin{Lem}\label{lem_extension}
The extended input does not affect the optimal cost or the optimal solutions. We can then assume without loss of generality that an optimal solution $\OPT$ consists of $k$-sets $\{S^*_1, S^*_2, \ldots, S^*_L\}$ and is disjoint, that is, $S^*_i \cap S^*_j=\emptyset$ for all $i$ and $j$ and $|S^*_i|=k$ for all $i$.
\end{Lem}

\begin{figure}[h!]
\centering
\includegraphics[height=2in]{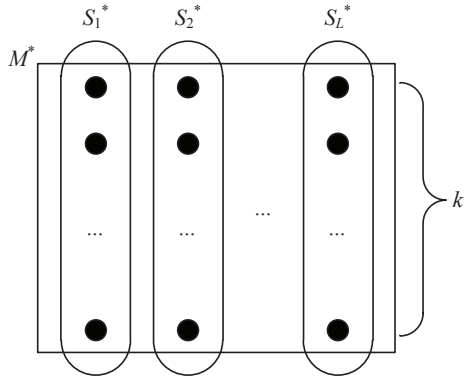}
\caption{The matrix $M^*$ representation \cite{Hassin.Levin.SICOMP05} of elements for the minimum $k$-set cover problem, containing exactly $k$ rows.}\label{fig:matrix}
\end{figure}

Thus, an optimal solution can be represented as a matrix $M^*$ of elements, as plotted in Figure~\ref{fig:matrix}, where column $i$ corresponds to the elements in $S^*_i$. Note that there are exactly $k$ rows in $M^*$, since each set in an optimal solution contains exactly $k$ elements. For an element $e$, we denote $M^*(e)$ as the column to which $e$ belongs, that is, the set in $\OPT$ that contains $e$. We also denote $w^*(e)$ as the cost of $M^*(e)$, and $N(e)$ as the number of uncovered elements in column $M^*(e)$ at the time at which $e$ is covered.

\subsection{SEIP ratio for the (unbounded) MSCP}

In Theorem \ref{them_setcover} we show that SEIP achieves an $H_k$-approximation ratio, where $k$ is the cardinality of the largest set. It has been proved that SEMO \cite{Friedrich.etal.GECCO07} achieves an $H_n$-approximation ratio for the MSCP, which is known as the asymptotic lower bound for the problem. The theorem confirms that SEIP can simulate multi-objective EAs in terms of the solution retained, so that SEIP is able to achieve the approximation ratio obtained by multi-objective EAs.

\begin{Them}\label{them_setcover}
Given an MSCP $(n, \bm w, C, U)$ where $|C|=m$, GSEIP finds an $H_k$-approximate solution in expected time $O(mn^2)$, where $k$ is the size of the largest set in $C$.
\end{Them}

The theorem is proved by simulating the greedy algorithm and using the property of the greedy algorithm as in Lemma \ref{lem_greedyratio}, which is derived from \cite{cheatal.mos79}.

\begin{Lem}\label{lem_greedyratio}
Given an MSCP $(n, \bm w, C, U)$ and an arbitrary partial solution $\bm x$, let $\hat{S} = \mathop{\arg\min}_S r_S$ with respect to $\bm x$. For every element $e\in \hat{S}$, there exists a set $M^*(e)$ of an optimal solution that covers $e$, and it holds that,
    \begin{align}
        price(e) \leq \frac{w(M^*(e))}{| M^*(e) - R(\bm x^{cur})|}.
    \end{align}
\end{Lem}

\begin{myproofd}{of Theorem \ref{them_setcover}}
    We find a non-negligible path following the greedy rule; given the current partial solution, add the set $\hat{S}$ with minimum $r_{\hat{S}}$, as in Algorithm \ref{algo_greedy}.

   We denote $\bm x^{cur}$ as the current solution to be operated on. We find $\bm y^+ = \{\hat{S}\}$ where the set $\hat{S}$ minimizes $r_{\hat{S}}$ with respect to $\bm x^{cur}$. Let $\bm y^- = \emptyset$. Thus, we have $|\bm y^+|+|\bm y^-|=1$ and $|\mu(\bm x\cup\bm y^+ - \bm y^-)| \geq |\mu(\bm x)|+1$ for partial solution $\bm x$.

By Lemma \ref{lem_greedyratio}, for all $e\in \hat{S}-R(\bm x^{cur})$, there exists a set $M^*(e)$ of an optimal solution that covers $e$, and suppose
    \begin{align}
        price(e) \leq \frac{w(M^*(e))}{| M^*(e) - R(\bm x^{cur})|}.
    \end{align}
    In the worst case, $|\hat{S}-R(\bm x^{cur})|=1$, that is, the added set only covers one uncovered element $e$. In this case, according to the definition of $price(e)$, we have $w(\hat{S})= price(e)$. We then have
    \begin{align}
        f(\bm y^+-\bm y^-\mid \bm x^{cur}) & = f(\bm x^{cur} \cup \bm y^+) - f(\bm x^{cur}) \\
        & = w(\hat{S}) \leq \frac{1}{| M^*(e) - R(\bm x^{cur})|} \cdot w(M^*(e)).
    \end{align}
    Thus, we find a non-negligible path with gap 1 and sum of ratios
    \begin{align}
        \sum_{e} \frac{1}{| M^*(e) - R(\bm x^{cur}(e))|} \cdot w(M^*(e)) & =
        \sum_{j=1}^{|\OPT|} \sum_{e\in S_j^*} \frac{1}{| M^*(e) - R(\bm x^{cur}(e))|} \cdot w(M^*(e)) \\
        & =  \sum_{j=1}^{|\OPT|} \sum_{e\in S_j^*} \frac{1}{| M^*(e) - R(\bm x^{cur}(e))|} \cdot w(S_j^*)\\
        & \leq \sum_{j=1}^{|\OPT|}  w(S_j^*)  \cdot \sum_{i=1}^k \frac{1}{| M^*(e) - R(\bm x^{cur}(e))|} \\
        & = \sum_{j=1}^{|\OPT|}  w(S_j^*) \cdot H_k \\
        & = H_k \cdot OPT,
    \end{align}
    where $\bm x^{cur}(e)$ denotes the partial solution that will cover $e$ in its next step, and $k$ is the size of the largest set in $C$.

By Theorem \ref{them_gsel}, GSEIP finds an $H_k$-approximate solution. Note that the isolation function maps to at most $n$ isolations, the non-negligible path has a constant gap of 1, and the solution is encoded in an $m$-length binary vector; thus, GSEIP takes expected time $O(mn^2)$.
\end{myproofd}

Note that in the proof of Theorem \ref{them_setcover}, with respect to $\bm x^{cur}$, we find $|\bm y^-|=0$ and $|\bm y^+|=1$. Thus, the proof can be adapted to LSEIP directly, as in Theorem \ref{them_setcover_l}.

\begin{Them}\label{them_setcover_l}
Given an MSCP $(n, \bm w, C, U)$ where $|C|=m$, LSEIP finds an $H_k$-approximate solution in expected $O(mn^2)$ time, where $k$ is the size of the largest set in $C$.
\end{Them}

\subsection{Ratio for GSEIP for the minimum $k$-set cover problem}

In this section, we prove in Theorem \ref{them_ksetcover} that GSEIP achieves the same $(H_k-\frac{k-1}{8k^9})$-approximation ratio as GAWW (Algorithm \ref{algo_gaww}) \cite{Hassin.Levin.SICOMP05}, the current best algorithm for the minimum $k$-set cover problem. This result reveals that when the problem is bounded, which is very likely in real-world situations, GSEIP can yield a better approximation ratio than in the unbounded situation. Since the greedy algorithm cannot achieve an approximation ratio lower than $H_n$, the result also implies that GSEIP has essential non-greedy behavior for approximations.

\begin{Them}\label{them_ksetcover}
Given a minimum $k$-set cover problem $(n, \bm w, C, U,k)$, where $|C|=m$, we denote $R(\bm x)=\cup_{S\in \bm x} S$. GSEIP using $\mu(\bm x)=R(\bm x)$ finds an $(H_k-\frac{k-1}{8k^9})$-approximate solution in expected time $O(m^{k+1}n^2)$.
\end{Them}

When applying the GAWW rule to select sets, we use Lemmas \ref{lem_gaww_disjoint} and \ref{lem_gaww_intersect}. Owing to the assignments of $price(\cdot)$, the total price of elements covered by $X$ equals the cost of $X$. We say a set $S$ is \emph{last-covered} if $\forall e\in S:N(e)=0$.
\begin{Lem}[Lemma 4 of \cite{Hassin.Levin.SICOMP05}]\label{lem_gaww_disjoint}
In each step of GAWW, its partial solution $H$ is disjoint.
\end{Lem}
\begin{Lem}[Lemma 5 of \cite{Hassin.Levin.SICOMP05}]\label{lem_gaww_intersect}
In each step of GAWW, we can assume without loss of generality that the set added to $X$ (i.e., $S^*$ or $\tilde{Q}$) has no more than one element in common with every set in $\OPT$.
\end{Lem}

Lemmas \ref{lem_gaww_2} to \ref{lem_gaww_1c} are derived from Lemmas 8, 9 and 10 in \cite{Hassin.Levin.SICOMP05} and are required for calculating the weights of sets that GSEIP selects following the GAWW rule.

\begin{Lem}\label{lem_gaww_2}
    Given a minimum $k$-set cover problem $(n, \bm w, C, U,k)$ and an arbitrary partial solution $\bm x$, if the GAWW rule uses a withdrawal step to add sets $\tilde{Q}=\{S_1, S_2, \ldots, S_l\}$ to $\bm x$ and withdrawal $\tilde{S}\in \bm x$, denoting $R=\cup_{S\in \tilde{Q}} S$, then
    \begin{align}
        \sum_{S\in \tilde{Q}} w(S) \leq w(\tilde{S}) + \sum_{{e \in R - \tilde{S}}} \frac{w^*(e)}{N(e)+1} - \sum_{{e \in R - \tilde{S}}} \frac{w^*(e)}{k^4}.
    \end{align}
\end{Lem}

\begin{Lem}\label{lem_gaww_1a}
Given a minimum $k$-set cover problem $(n, \bm w, C, U,k)$ and an arbitrary partial solution $\bm x$, if the GAWW rule selects a set $S$ that is not last-covered to add to $\bm x$ (greedy step), and there is an element $e'\in S$ such that
    \begin{align}
        price(e') < w^*(e') (\frac{1}{L_i+1}-\frac{1}{4k^5}),
    \end{align}
    then
    \begin{align}
        w(S) \leq \sum\limits_{e\in S} \frac{w^*(e)}{N(e)+1} - \sum\limits_{e\in S} \frac{w^*(e)}{8k^8}.
    \end{align}
\end{Lem}

\begin{Lem}\label{lem_gaww_1b}
Given a minimum $k$-set cover problem $(n, \bm w, C, U,k)$ and an arbitrary partial solution $\bm x$, if the GAWW rule selects a set $S$ that is not last-covered to add to $\bm x$ (greedy step), followed by another set $S'$, and for all elements $e\in S$ for which
    \begin{align}
        price(e) \geq w^*(e') \left(\frac{1}{h_i+1}-\frac{1}{4k^5}\right),
    \end{align}
    then
    \begin{align}
        w(S)+w(S') \leq \sum_{e\in S} \frac{w^*(e)}{N(e)+1} - \sum_{e\in S} \frac{w^*(e)}{8k^8}.
    \end{align}
\end{Lem}

\begin{Lem}\label{lem_gaww_1c}
Given a minimum $k$-set cover problem $(n, \bm w, C, U,k)$ and an arbitrary partial solution $\bm x$, if the GAWW rule selects a set $S$ that is last-covered to add to $\bm x$ (greedy step), then
    \begin{align}
        w(S) \leq \sum\nolimits_{e\in S} \frac{w^*(e)}{N(e)+1}.
    \end{align}
\end{Lem}

\begin{myproofd}{of Theorem \ref{them_ksetcover}}
    We find a path of isolations following the GAWW rule. Note that there are at most $n+1$ isolations.

For every $\bm x^{cur}$ belonging to an isolation $|\mu(\bm x^{cur})|$ on the path, if the GAWW rule selects sets satisfying Lemmas \ref{lem_gaww_2}, \ref{lem_gaww_1a}, \ref{lem_gaww_1b} and \ref{lem_gaww_1c}, we find $\bm y^+$ containing the sets added and $\bm y^-$ containing the set withdrawn. Thus, $|\bm y^+|+|\bm y^-|\leq k+1$.

Since the GAWW rule covers at least one uncovered element, we have $|\mu(\bm x^{cur}\cup \bm y^+ - \bm y^-)|>|\mu(\bm x^{cur})|$.

As long as no last-covered set is included, by Lemmas \ref{lem_gaww_2}, \ref{lem_gaww_1a} and \ref{lem_gaww_1b}, the partial ratio of $f(\bm y^+-\bm y^-\mid \bm x^{cur}) = w(\bm y^+)-w(\bm y^-)$ is upper-bounded as
    \begin{align}
        w(\bm y^+)-w(\bm y^-) \leq \sum\limits_{e\in (\cup_{S\in \bm y^+}) - (\cup_{S\in \bm y^-})} \left(\frac{w^*(e)}{N(e)+1} - \frac{w^*(e)}{8k^8}\right);
    \end{align}
    otherwise, when the GAWW rule selects a last-covered set using a greedy step, by Lemma \ref{lem_gaww_1c}, noting $\bm y^-=\emptyset$,
    \begin{align}
        w(\bm y^+)-w(\bm y^-) \leq \sum_{e\in (\cup_{S\in \bm y^+}) - (\cup_{S\in \bm y^-})} \left(\frac{w^*(e)}{N(e)+1}\right).
    \end{align}
    In the worst case, there are $1/k$ among all the elements that are last-covered. Therefore, we find a non-negligible path.

    By Theorem \ref{them_gsel}, GSEIP finds a solution with approximation ratio
    \begin{align}
        \frac{1}{OPT}\left(\sum\nolimits_{i=1}^{k} \frac{OPT}{i} - \frac{k-1}{k} \frac{OPT}{8k^8}\right) = H_k - \frac{k-1}{8k^9},
    \end{align}
    in expected time $O(m^{k+1}n^2)$.
\end{myproofd}

\subsection{Comparison of the greedy algorithm, LSEIP and GSEIP}

We assess the greedy algorithm, LSEIP and GSEIP for a subclass of the minimum $k$-set cover problem, denoted as problem $I$. We show in Propositions \ref{prop_worst-greedy} to \ref{prop_worst-gseip} that LSEIP and GSEIP can overcome the difficulty that limit the greedy algorithm, and thus yield better approximation ratios.

\begin{figure}[h!]
\centering
\includegraphics[height=2in]{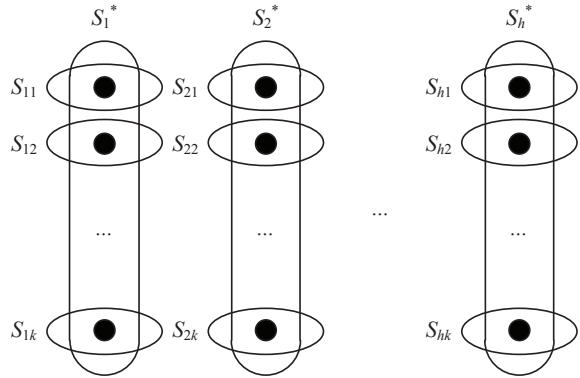}
\caption{Subclass $I$ of the minimum $k$-set cover problem.}\label{fig:worst-k}
\end{figure}

Problem $I$ is a minimum $k$-set cover problem $(n, \bm w, C, U)$ constructed as follows. Note that $n=k L$ for some integer $L$. The optimal solution consists of $L$ non-overlapping sets $\{S^*_i\}_{i=1}^L$, each of which contains $k$ elements in $U$. Imagine that elements in each $S^*_i$ are ordered and let $S_{ij}$ contain only the $j$th element of set $S^*_i$. The collection of sets $C$ consists of all sets of $S^*_i$ and $S_{ij}$. Assign each $S^*_i$ weight $1+\epsilon$ for some $\epsilon>0$, and assign each $S_{ij}$ weight $1/j$. Problem $I$ thus constructed is shown in Figure \ref{fig:worst-k}.

\begin{Prop}\label{prop_worst-greedy}
    Given an arbitrary value $\xi>0$, the approximation ratio of the greedy algorithm for problem $I$ is lower-bounded by $H_k-\xi$.
\end{Prop}
\begin{myproof}
Note that the greedy algorithm adds the set with the minimum cost to the solution at each step. On initialization, the cost of $S_i^*$ for all $i$ is $\frac{1+\epsilon}{k}$, which is higher than the cost $\frac{1}{k}$ of $S_{ik}$, the smallest cost. Thus, $S_{ik}$ for all $i$ will be added to the solution. Then the cost of $S_i^*$ for all $i$ is $\frac{1+\epsilon}{k-1}$, which is higher than $\frac{1}{k-1}$ for $S_{i(k-1)}$. The greedy algorithm continues to choose a non-optimal set. Finally, all sets $S_{ij}$ will be added to the solution. The approximation ratio is therefore $H_k - H_k\cdot \frac{\epsilon}{1+\epsilon}$. Let $\xi = H_k\cdot\frac{\epsilon}{1+\epsilon}$. $\xi$ can be an arbitrarily small positive value as $\epsilon$ can be arbitrarily small.
\end{myproof}

\begin{Prop}\label{prop_worst-lseip}
With probability of at least $1-\frac{1}{k+1}$, LSEIP finds a $\left(H_k - \frac{k}{n}(H_k-1)\right)$-approximate solution for problem $I$ in $O\left((1+\frac{1}{k})n^3\right)$ expected steps.
\end{Prop}
\begin{myproof}
Let LSEIP run until there are $k$ elements uncovered by any solution identified. This takes $O(mn^2)$ expected steps by Theorem \ref{them_setcover_l}, which is $ O((1+\frac{1}{k})n^3)$ since $m=(1+\frac{1}{k})n$. The uncovered $k$ elements would be covered by sets in $\{S_i^*\}$. In the worst case, we assume the $k$ elements are in one set $S_{\hat{i}}^*$ for $\hat{i}$.

Given a solution to be operated on, $S_{\hat{i}}^*$ is selected with probability $\frac{1}{m}$; set $S_{\hat{i}j}$ for $j\in\{1,2,\ldots,k\}$ is selected with probability $\frac{k}{m}$; with the remaining probability, no more elements will be covered and we return to the solution. Therefore, to cover one of the $k$ elements, $S_{\hat{i}}^*$ is selected with probability $\frac{1}{k+1}$; and set $S_{\hat{i}j}$ for $j\in\{1,2,\ldots,k\}$ is selected with probability $\frac{k}{k+1}$.

Suppose one set $S_{\hat{i}\hat{j}}$ is selected. To cover one more uncovered element, $S_{\hat{i}}^*$ is selected with probability $\frac{1}{k}$; $S_{\hat{i}j}$ for $j\in\{1,2,\ldots, \hat{j},\ldots, k\}$ is selected with probability $\frac{k-1}{k}$. Therefore, the set $S_{\hat{i}}^*$ is selected with probability
    \begin{align}
        &\frac{1}{k+1}  + \frac{1}{k}(1-\frac{1}{k+1}) + \ldots + \frac{1}{2} \prod_{i=3}^{k+1}(1-\frac{1}{i})\\
        & = \frac{1}{k+1} \cdot \frac{k+1}{k+1}+ \frac{1}{k}\cdot \frac{k}{k+1} + \ldots + \frac{1}{2} \cdot \frac{2}{k+1}\\
        & = \frac{k}{k+1} = 1 - \frac{1}{k+1},
    \end{align}
    and within $O((k+1)mn)$ steps, which is overwhelmed by $O(mn^2)$.

In the worst case, only one set $S_{\hat{i}}^*$ of an optimal solution is selected for the feasible solution. Thus, the approximation ratio is
    \begin{align}
        \frac{(L-1)H_k+1+\epsilon}{L(1+\epsilon)} \leq \frac{L-1}{L} H_k + \frac{1}{L} = H_k - \frac{1}{L}(H_k-1) =  H_k - \frac{k}{n}(H_k-1).
    \end{align}
\end{myproof}

\begin{Prop}\label{prop_worst-gseip}
GSEIP finds the optimal solution for problem $I$ in $O(\frac{1}{k}\cdot(1+\frac{1}{k})^{k+1}\cdot n^{k+3} )$ expected steps.
\end{Prop}

\begin{myproof}
Let GSEIP run until a feasible solution is found, which takes $O(mn^2)$ expected steps according to Theorem \ref{them_setcover}. In the worst case, suppose the feasible solution found consists of all sets $S_{ij}$, so that the approximation ratio is $H_k-\xi$ for an arbitrary small value $\xi>0$.

Keep GSEIP running until an optimal solution is found. GSEIP chooses to operate on the feasible solution with probability of at least $\frac{1}{n}$ as there are $n$ isolations, which implies there are $O(n)$ steps.

Operating on the feasible solution, for some $i$, GSEIP uses its mutation operator to replace sets $S_{ij}$ for all $j$ and with $S_i^*$ with probability $(\frac{m-1}{m})^{m-k-1} (\frac{1}{m})^{k+1}$, since $k+1$ bits in the solution are to be flipped, which implies there are $O(m^{k+1})$ steps. Once sets $S_{ij}$ for all $j$ are replaced with $S_i^*$, the partial ratio (approximation ratio) decreases and thus the mutated solution is retained.

Therefore, GSEIP takes $O(mn^2+ nm^{k+1}L)$ steps in all; note that $O(mn^2+ nm^{k+1}L) = O(mn^2+ n^2m^{k+1}/k) = O(\frac{1}{k}\cdot(1+\frac{1}{k})^{k+1}\cdot n^{k+3} )$, since $m = n(1+\frac{1}{k})$.
\end{myproof}

For GSEIP, we can also derive Proposition \ref{prop_worst-progress}, proof of which is similar to that for Proposition \ref{prop_worst-gseip}.

\begin{Prop}\label{prop_worst-progress}
GSEIP finds an $\left(H_k - c\frac{k}{n}(H_k-1)\right)$-approximate solution for problem $I$ in $O(mn^2+ cn^2m^{k+1}/k)$ expected steps, for $c=0, 1, \ldots, \frac{n}{k}$.
\end{Prop}

This proposition illustrates an interesting property: compared with the greedy algorithm, whose performance cannot be improved given extra time, SEIP always seeks better approximate solutions. Users can allocate more running time in a trade-off for better solutions. However, the solution quality may not be improved by an EA over a long time for practical HP-hard problems. Thus, users should not assume any useful relationship between time allocated and the approximate ratio.

\section{Conclusion}

We studied the approximation performance of an EA framework. SEIP introduces an isolation function to manage competition among solutions, which can be configured as a single- or multi-objective EA. We analyzed the approximation performance of SEIP using the partial ratio and obtained a general characterization of the SEIP approximation behavior. Our analysis confirms that SEIP can achieve a guaranteed approximate solution: it tries to optimize the partial ratio of solutions in every isolation by finding good partial solutions, then these partial solutions form a feasible solution with a good approximation ratio.

We studied the performance of SEIP for the MSCP, which is NP-hard. Previous studies \cite{Friedrich.etal.GECCO07,Oliveto.etal.TEC09} showed that (1+1)-EA is not a good solver for MSCP, while a multi-objective EA achieves a similar approximation ratio to that of the greedy algorithm. Our analysis extends previous work to show that for the unbounded MSCP, SEIP achieves an $H_k$-approximation ratio (where $H_k$ is the harmonic number of the cardinality of the largest set), the asymptotic lower bound. For the minimum $k$-set cover problem, it achieves an $(H_k-\frac{k-1}{8k^9})$-approximation ratio, the current best-achievable result \cite{Hassin.Levin.SICOMP05}, which is beyond the ability of the greedy algorithm. Moreover, for a subclass of the MSCP, we show how SEIP with either one-bit or bit-wise mutation can overcome the difficulty that limits the greedy algorithm.

We discussed some advantages and limitations of SEIP for approximations. To prove the SEIP approximation ratio for the MCSP problem, we used SEIP to simulate the greedy algorithm. Since the greedy algorithm is a general scheme for approximations and has been analyzed for many problems, SEIP analysis can be extended to cases for which the greedy algorithm has been applied, and therefore we can easily show that SEIP is a $\frac{1}{k}$-approximation algorithm for $k$-extensible systems \cite{mestre.esa06}, including $b$-matching, maximum profit scheduling and maximum asymmetric TSP problems. Moreover, to prove the approximation ratio of SEIP for the minimum $k$-set cover problem, we used SEIP to simulate the GAWW algorithm, which implies that SEIP also has extra behaviors that provide opportunities to exceed the greedy algorithm. However, limitations of SEIP are found from Theorem \ref{them_gsel}. There are some situations in which SEIP may fail. SEIP is required to flip a number of bits at a time, which depends on $n$, to achieve a good solution. In this situation, to flip the required number of bits, one-bit mutation is limited since it only flips one bit at a time, and bit-wise mutation is also limited as it requires exponential time. Further designs for elegant mutation operators may be possible. However, since the solution space has size $2^n$, any mutation has to assign an exponentially small probability to some distant mutations. Once these distant mutations are required, exponential time is required. Another view of this limitation is that since the mutation operator only allows limited steps, SEIP may also over-optimize partial ratios, just like the greedy algorithm over-optimizes each step.

Our theoretical analysis suggests that EAs can achieve solutions with guaranteed performance. We believe that guaranteed and better approximation performance can be achieved by better EA design in the future.

\section*{Acknowledgements}
We thank Dr. Per Kristian Lehre and Dr. Yitong Yin for helpful discussions. This work was partly supported by the National Fundamental Research Program of China (2010CB327903), the National Science Foundation of China (61073097, 61021062), EPSRC (UK) (EP/I010297/1), and an EU FP7-PEOPLE-2009-IRSES project under Nature Inspired Computation and its Applications (NICaiA) (247619).

\bibliography{ref}
\bibliographystyle{abbrvnat}

\end{document}